\documentclass[10pt,twocolumn,letterpaper]{article}

\usepackage{iccv}
\usepackage{times}
\usepackage{epsfig}
\usepackage{graphicx}
\usepackage{amsmath}
\usepackage{amssymb}

\usepackage{textcomp}
\usepackage{listings}
\usepackage{caption}
\usepackage{subcaption}
\usepackage{threeparttable}
\usepackage{multirow}
\usepackage{array}
\usepackage{booktabs}

\makeatletter
\renewcommand{\paragraph}{%
  \@startsection{paragraph}{4}%
  {\z@}{1.1ex \@plus 1ex \@minus .2ex}{-1em}%
  {\normalfont\normalsize\bfseries}%
}
\makeatother
\usepackage[pagebackref=true,breaklinks=true,letterpaper=true,colorlinks,bookmarks=false]{hyperref}

\iccvfinalcopy

\ificcvfinal\fi

\begin{document}

\title{In-Place Scene Labelling and Understanding with Implicit Scene Representation}

\author{Shuaifeng Zhi, Tristan Laidlow, Stefan Leutenegger, Andrew J. Davison\\
Dyson Robotics Laboratory at Imperial College\\
Department of Computing, Imperial College London, UK\\
{\tt\small \{s.zhi17, t.laidlow15, s.leutenegger, a.davison\}@imperial.ac.uk}}

\maketitle
\ificcvfinal\thispagestyle{empty}\fi

\begin{abstract}
\vspace{-2mm}
Semantic labelling is highly correlated with geometry and radiance reconstruction, as scene entities with similar shape and appearance are more likely to come from similar classes. Recent implicit neural reconstruction techniques are appealing as they do not require prior training data, but the same fully self-supervised approach is not possible for semantics because labels are human-defined properties.

We extend neural radiance fields (NeRF) to jointly encode semantics with appearance and geometry, so that complete and accurate 2D semantic labels can be achieved using a small amount of in-place annotations specific to the scene. The intrinsic multi-view consistency and smoothness of NeRF benefit semantics by enabling sparse labels to efficiently propagate. We show the benefit of this approach when labels are either sparse or very noisy in room-scale scenes. 
We demonstrate its advantageous properties in various interesting applications such as an efficient scene labelling tool, novel semantic view synthesis, label denoising, super-resolution, label interpolation and multi-view semantic label fusion in visual semantic mapping systems.
\vspace{-6mm}
\end{abstract}

\section{Introduction}
\vspace{-1mm}
Enabling intelligent agents, such as indoor mobile robots, to plan context-sensitive actions in their environment requires both a geometric and semantic understanding of the scene.
Machine learning methods have proven to be valuable in both geometric and semantic prediction tasks, but the performance of these methods suffers when the distribution of the training data does not match the scenes observed at test-time. Though the issue can be mitigated by gathering costly annotated data or semi-supervised learning, it is not always feasible in open-set scenarios with various known and unknown classes.
For this reason, it is advantageous to have methods that can self-supervise. In particular, there has been recent success in using scene-specific methods (e.g. NeRF \cite{Mildenhall:etal:ECCV2020}) that represent the shape and radiance of a single scene with a neural network trained from scratch using only images and associated camera poses.

\begin{figure}[!t]
\centering
\includegraphics[width =1\linewidth]{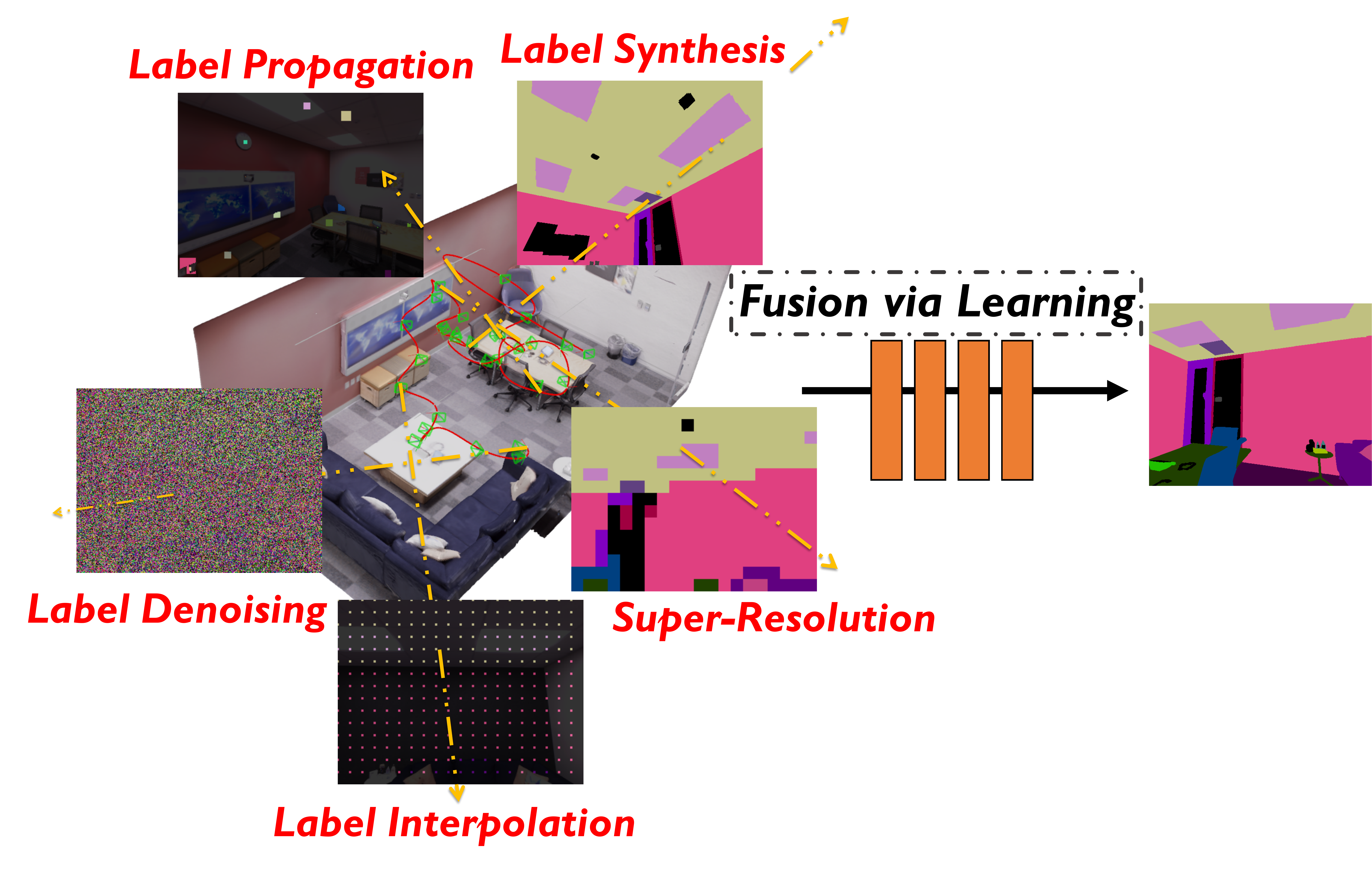}
\caption{Neural radiance fields (NeRF) jointly encoding appearance and geometry contain strong priors for segmentation and clustering. 
We build upon this to create a scene-specific 3D semantic representation, Semantic-NeRF, and show that it can be efficiently learned with in-place supervision to perform various potential applications.} 
\label{fig:teaser}
\vspace{1mm} \hrule \vspace{-5mm}
\end{figure}

Semantic scene understanding means attaching class labels to a geometric model.
The tasks of estimating the geometry of a scene and predicting its semantic labels are strongly related, as parts of a scene that have similar shape are more likely to belong to the same semantic category than those which differ greatly.
This has been shown in work on multi-task learning \cite{Shikun:etal:CVPR2019, Zamir:etal:CVPR2018}, where networks that simultaneously predict both shape and semantics perform better than when the tasks are tackled separately.

Unlike scene geometry, however, semantic classes are a human-defined concept and it is not possible to semantically label a novel scene in a purely self-supervised manner.
The best that could be achieved would be to cluster self-similar structures of a scene into categories; but some labelling would always be needed to associate these clusters with human-defined semantic classes.

In this paper, we show how to design a scene-specific network for joint geometric and semantic prediction and train it on images from a single scene with only weak semantic supervision (and no geometric supervision). Because our single network must generate both geometry and semantics, the correlation between these tasks means that semantics prediction can benefit from the smoothness, coherence and self-similarity learned by self-supervision for geometry. In addition, multi-view consistency is inherent to the training process and enables the network to produce accurate semantic labels of the scene, including for views that are substantially different from any in the input set. 

Our system takes as input a set of RGB images with associated known camera poses. We also supply some partial or noisy semantic labels for the images, such as ground truth labels for a small fraction of the images, or noisy or coarse label maps for a higher number of images. We train our network to jointly produce implicit 3D representations of both the geometry and semantics for the whole scene.

We evaluate our system both quantitatively and qualitatively on scenes from the Replica dataset \cite{Straub:etal:ARXIV2019}, and qualitatively on real-world scenes from the ScanNet dataset \cite{Dai:etal:CVPR2017}. Generating dense semantic labels for a whole scene from partial or noisy input labels is important for practical applications, like when a robot encounters a new scene and either only a small amount of in-situ labelling is feasible, or only an imperfect single-view network is available. 
\vspace{-1mm}
\section{Related Work}

Most existing 3D semantic mapping and understanding systems work by attaching (fused) semantic labels to a 3D geometric representation created by a standard reconstruction method.
For example, \cite{Hermans:etal:ICRA2014} uses point clouds, \cite{McCormac:etal:ICRA2017} and \cite{Runz:etal:ISMAR2018} use surfels, \cite{Narita:etal:IROS2019} uses voxels, and \cite{McCormac:etal:3DV2018} uses signed distance fields.
These classical 3D geometric representations are all limited in their ability to efficiently represent fine details in complex topologies.
Volumetric representations, for example, have a convenient structure for parallel processing or use with convolutional neural networks, but suffer from large memory requirements due to discretisation that ultimately limits the resolution it can represent.

\subsection{Code-Based Representations}

To help overcome these limitations, many learning-based representations have been developed.
Code-based representations, for example, use the latent code of an autoencoder as a compact representation of the scene.
Generative Query Networks (GQN) \cite{Eslami:etal:Science2018} can represent simple 3D scenes using a latent scene representation vector.
CodeSLAM \cite{Bloesch:etal:CVPR2018} uses a compact and optimisable latent code to represent dense scene geometry in a view-based visual odometry system.
The approach of CodeSLAM was extended by SceneCode \cite{Zhi:etal:CVPR2019} to include semantics.
SceneCode was able to make inference-time refinements of network predictions by optimising for photometric and semantic label consistency between multiple frames.
However, although trained with depth maps, SceneCode was still a view-based representation and lacked true awareness of the 3D geometry.

\subsection{Implicit 3D Representations}

There has been much promising recent work on using neural implicit scene representations.
As these are continuous representations, they can easily handle complicated topologies and do not suffer from discretisation error, with the actual representative resolution depending on the capacity of the neural network used.
The Scene Representation Network (SRN) \cite{Vincent:2019:NIPS2019} was one of the first methods to use a multi-layer perceptron (MLP) as the neural representation of a learned scene given a collection of images and associated poses.
DeepSDF \cite{Park:etal:CVPR2019} and DIST \cite{Liu:etal:CVPR2020} used deep decoders to learn implicit signed distance functions of various shape instances of the same class, and Occupancy Networks \cite{Mescheder:etal:CVPR2019, Peng:etal:ECCV2020} learned an implicit 3D occupancy function for shapes or large scale scenes given 3D supervision.

Kohli \etal \cite{Amit:etal:3DV2020} also proposed to learn a joint implicit representation of appearance and semantics for 3D shapes on top of an SRN using a linear segmentation renderer. After being trained in a two-step semi-supervised manner, the network can synthesise novel view semantic labels from either colour or semantic observations.

The methods mentioned above involve extensive pre-training on collections of data to learn priors about the shapes or scenes they are used to represent. Although promising generalisation capability has been shown across different instances or scenes, it is not always possible to get adequate data for various unseen environments. The alternative is  a scene-specific representation which requires minimum in-place labelling effort.
 
NeRF \cite{Mildenhall:etal:ECCV2020} and other systems based on it \cite{Zhang:etal:ARXIV2020NeRF++, Martin:etal:CVPR2021, Trevithick:Yang:arxiv2020, Srinivasan:etal:CVPR2021} use MLPs to overfit input from a single bounded scene and act as an implicit volumetric representation for realistic view-synthesis.
In this paper, we treat NeRF as a powerful scene-specific 3D implicit representation, and extend it to include semantic representation which can be efficiently learned from sparse or noisy annotations (Figure \ref{fig:teaser}).

\vspace{-1mm}
\section{Method}
\subsection{Preliminaries}

Given multiple images of a static scene with known camera intrinsics and extrinsics, NeRF \cite{Mildenhall:etal:ECCV2020} uses MLPs to implicitly represent the continuous 3D scene density $\sigma$ and colour $\mathbf{c} = (r,g,b)$ as a function of continuous 5D input vectors of spatial coordinates $\mathbf{x}=(x,y,z)$ and viewing directions $\mathbf{d}=(\theta, \phi)$. Specifically, $\sigma(\mathbf{x})$ is designed to be a function of only 3D position while the radiance $\mathbf{c(\mathbf{x}, \mathbf{d})}$ is a function of both 3D position and viewing direction.

To compute the colour of a single pixel, NeRF \cite{Mildenhall:etal:ECCV2020} approximates volume rendering by numerical quadrature with hierarchical stratified sampling. 
Within one hierarchy, if $ \mathbf{r}(t) = \mathbf{o}+t\mathbf{d}$ is the ray emitted from the centre of projection of camera space through a given pixel, traversing between near and far bounds ($t_n$ and $t_f$), then for selected $K$ random quadrature points $\{ t_k \}_{k=1}^{K}$ between $t_n$ and $t_f$, the approximated expected colour is given by:
\vspace{-3mm}
\begin{align}
    \hat{{\mathbf{C}}}(\mathbf{r}) =&
        \sum_{k=1}^{K} \hat{T}(t_k) \, \alpha \left( {\sigma(t_k) \delta_{k}}\right) \mathbf{c}(t_k) \, , \\
    \text{where} \quad \hat{T}(t_k) = &
        \exp \left(-\sum_{k'=1}^{k-1} \sigma(t_k) \delta_k \right) \, ,
\end{align}
where $\alpha \left({x}\right) = 1-\exp(-x)$, and $\delta_k = t_{k+1} - t_k$ is the distance between two adjacent quadrature sample points.

Given multi-view training images of the observed scene, NeRF uses stochastic gradient descent (SGD) to optimise $\sigma$ and $\mathbf{c}$ by minimising  photometric discrepancy.
\subsection{Semantic-NeRF} \label{sec:semantic-nerf}
\begin{figure}[t!]
\centering
\includegraphics[width = 0.95\linewidth]{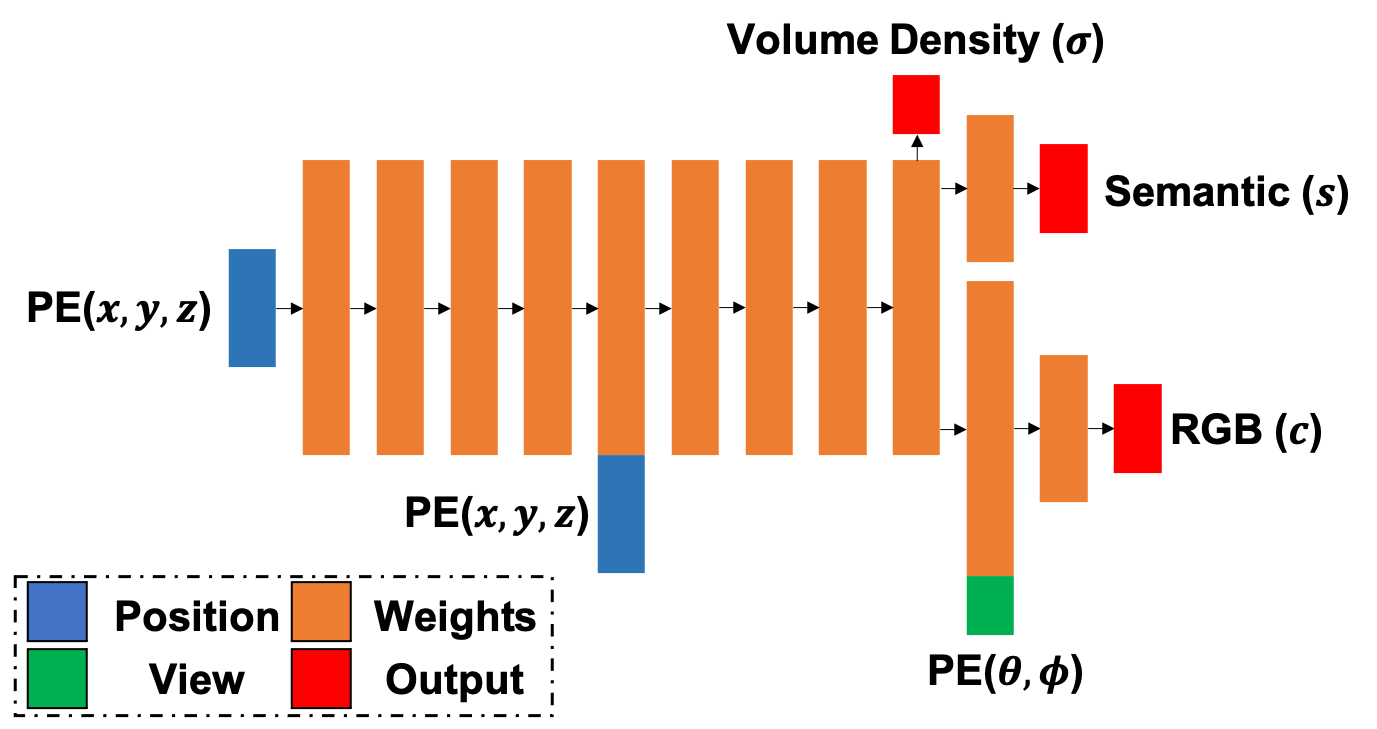}
\caption{Semantic-NeRF network architecture. 3D position ($x$, $y$, $z$) and viewing direction ($\theta$, $\phi$) are fed into the network after positional encoding (PE). Volume density $\sigma$ and semantic logits $\mathbf{s}$ are functions of 3D position while colours $\mathbf{c}$  additionally depend on viewing direction.}

\label{fig:nerf-architecture}
    \vspace{2mm} \hrule \vspace{-5mm}
\end{figure}

We now show how to extend NeRF to jointly encode appearance, geometry and semantics. As shown in Figure \ref{fig:nerf-architecture}, we augment the original NeRF by appending a segmentation renderer before injecting viewing directions into the MLP.

We formalise semantic segmentation as an inherently view-invariant function that maps only a world coordinate $\mathbf{x}$ to a distribution over $C$ semantic labels via pre-softmax semantic logits $\mathbf{s} (\mathbf{x})$:
\begin{equation}
\mathbf{c} = F_{\Theta}(\mathbf{x}, \mathbf{d}) \, , \quad \mathbf{s} = F_{\Theta}(\mathbf{x}) \, ,\label{eq:seg_renderer}
\end{equation}
where $F_{\Theta}$ represents the learned MLPs.

The approximated expected semantic logits $\hat{\mathbf{S}}(\mathbf{r})$ of a given pixel in the image plane can be written as:
\begin{align}
    \hat{{\mathbf{S}}}(\mathbf{r}) =&
        \sum_{k=1}^{K} \hat{T}(t_k) \, \alpha \left( {\sigma(t_k) \delta_{k}}\right) \mathbf{s}(t_k) \, , \\
    \text{where} \quad \hat{T}(t_k) = &
        \exp \left(-\sum_{k'=1}^{k-1} \sigma(t_k) \delta_k \right) \, ,
\end{align}
with $\alpha \left({x}\right) = 1-\exp(-x)$ and $\delta_k = t_{k+1} - t_k$ is the distance between adjacent sample points. Semantic logits can then be transformed into multi-class probabilities through a softmax normalisation layer.

\subsection{Network Training}
We train the whole network from scratch under photometric loss $L_p$ and semantic  loss $L_s$:
\begin{align}
L_p & =\sum_{\mathbf{r}\in \mathcal{R}}\left[{\left\lVert \hat{\mathbf{C}}_c(\mathbf{r}) - \mathbf{C}(\mathbf{r}) \right\rVert}_2^2+{\left\lVert \hat{\mathbf{C}}_f(\mathbf{r}) - \mathbf{C}(\mathbf{r}) \right\rVert}_2^2 \right], \label{eqn:loss_photo} 
\\
L_s & = - \sum_{\mathbf{r}\in \mathcal{R}}\left[
\sum_{l=1}^{L} p^{l}(\mathbf{r})\log \hat{p}_{c}^{l}(\mathbf{r})+
\sum_{l=1}^{L} p^{l}(\mathbf{r})\log \hat{p}_{f}^{l}(\mathbf{r})
\right], \label{eqn:loss_sem_1} 
\end{align}
where $\mathcal{R}$ are the sampled rays within a training batch, and $\mathbf{C}(\mathbf{r})$, $\hat{\mathbf{C}}_c(\mathbf{r})$ and $\hat{\mathbf{C}}_f(\mathbf{r})$ are the ground truth, coarse volume predicted and fine volume predicted RGB colours for ray $\mathbf{r}$, respectively. Similarly, ${p}^{l}$, $\hat{p}_{c}^{l}$ and $\hat{p}_{f}^{l}$ are the multi-class semantic probability at class $l$ of the ground truth map, coarse volume and fine volume predictions for ray $\mathbf{r}$, respectively.
$L_s$ is chosen as a multi-class cross-entropy loss to encourage the rendered semantic labels to be consistent with the provided labels, whether these are ground-truth, noisy or partial observations.
Hence the total training loss $L$ is:
\begin{equation}
    L = L_p + \lambda L_s \, ,
\end{equation}
where $\lambda$ is the weight of the semantic loss and is set to 0.04 to balance the magnitude of both losses \cite{Amit:etal:3DV2020}. In practice we find that actual performance is not sensitive to $\lambda$ value and setting $\lambda$ to 1 gives us similar performance.
These photometric and semantic losses naturally encourage the network to generate multi-view consistent 2D renderings from the underlying joint representation.

\subsection{Implementation}

A scene-specific semantic representation is obtained by training the network from scratch for each scene individually. 
We use setup and hyper-parameters similar to \cite{Mildenhall:etal:ECCV2020}. Specifically, we use hierarchical volume sampling to jointly optimise coarse and fine networks, where the former provides importance sampling bias so that the latter can distribute more samples to positions likely to be visible. Positional encoding of length 10 and 4 \cite{Vaswani:etal:NIPS2017,Tancik:etal:NIPS2020} are applied to 3D positions and viewing directions, respectively. In addition, since we have no depth information, we set the bounds of ray sampling to 0.1m and 10m respectively across experiments without careful tuning to span indoor scenes. 

Training images are resized to 320x240 for all the experiments. We implement our model in PyTorch \cite{Adam:etal:NIPS2019} and train it on a single RTX2080-Ti GPU with 11GB memory. The batch size of rays is set to 1024 due to memory limitations. We train the neural network using the Adam optimiser \cite{Kingma:Ba:ICLR2015} with a learning rate of 5e-4 for 200,000 iterations.
\vspace{-2mm}
\section{Experiments and Applications}
\vspace{-2mm}
After training on colour images and semantic labels with associated poses, we obtain a scene-specific implicit 3D semantic representation. We evaluate its effectiveness quantitatively by projecting the 3D representation back into 2D image space where we have direct access to explicit ground truth data.
We aim to show the benefits and promising applications of efficiently learning such a joint 3D representation for semantic labelling and understanding. We kindly urge readers to inspect more qualitative results on project page: \url{https://shuaifengzhi.com/Semantic-NeRF/}.

\vspace{-2mm}
\subsection{Indoor Scene Datasets and Data Preparation}
\vspace{-2mm}
\paragraph{Replica}\label{sec:dataset-replica}

Replica \cite{Straub:etal:ARXIV2019} is a reconstruction-based 3D dataset of 18 high fidelity scenes with dense geometry, HDR textures and semantic annotations.  We use the Habitat simulator \cite{Manolis:etal:ICCV2019} to render RGB colour images, depth maps and semantic labels from randomly generated 6-DOF trajectories similar to hand-held camera motions. We follow the procedure from SceneNet RGB-D \cite{McCormac:etal:ICCV2017}, and lock the roll angle with the camera up-vector pointing along the y-axis.

We use the provided 88 semantic classes from Replica in scene-specific experiments and also manually map these labels to the popular NYUv2-13 definition \cite{Silberman:etal:ECCV2012, Eigen:etal:ICCV2015} in Section~\ref{subsubsec: mvsf} for multi-view label fusion, following the mapping convention from ScanNet \cite{Dai:etal:CVPR2017}.
For each Replica scene of rooms and offices, we render 900 images at resolution 640x480 using the default pin-hole camera model with 90 degree horizontal field of view. We sample every 5th frame from the sequence to compose the training set and also sample intermediate frames to make the test set.
\vspace{-2mm}
\paragraph{ScanNet}
ScanNet \cite{Dai:etal:CVPR2017} is a large-scale real-world indoor RGB-D video dataset of 2.5M views in 1513 scenes with rich annotations including semantic segmentation, camera poses and surface reconstructions. We train Semantic-NeRF on ScanNet scenes using only the provided colour images, camera poses and 2D semantic labels. The sequences in each scene are evenly sampled so that the total amount of training data is roughly 300 frames. During experiments we select several indoor room-scale scenes and train one Semantic-NeRF per scene using posed images and semantic labels from the NYUv2-40 definition. 
\vspace{-2mm}
\subsection{Semantic Neural Radiance Fields}\label{sec:S-NeRF}
\vspace{-1mm}
We check the influence of semantics on appearance and geometry by quantitatively computing the quality of rendered RGB images and depth maps on Replica scenes with and without semantic prediction enabled. Experiments show that there is no clear difference which suggests that the current network has the capacity to learn these tasks jointly. Note that we might expect that significant high quality semantic labelling information could feasibly improve reconstruction quality, but in this paper we are focused on how geometry can help semantics in the opposite situation where semantic labelling is sparse or noisy.

\begin{figure*}[!t]
\centering
\includegraphics[width=1\linewidth]{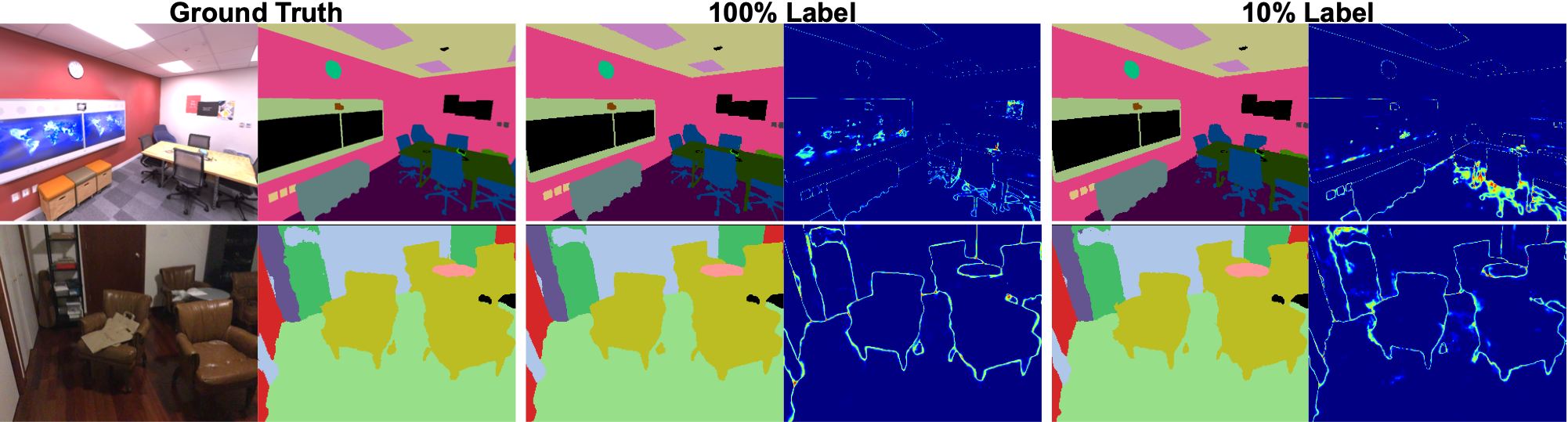}
\caption{Synthesised semantic labels at testing poses given 100\% and 10\% of ground truth labels during training. 
From left to right we show the ground truth colour and semantic images for reference, and rendered semantic labels and their information entropy given 100\% and 10\% supervision, respectively. Bright parts of the entropy map match well to object boundaries or ambiguous/unknown regions in the corresponding training set-up.
\label{fig:quali-sparse-vs}}
    \vspace{1mm} \hrule \vspace{-4mm}
\end{figure*}
\vspace{-2mm}
\subsection{Semantic View Synthesis with Sparse Labels}\label{sec:semantic-view-synthesis}
\vspace{-1.5mm}
We first train our semantic-NeRF framework for novel view semantic label synthesis using all available RGB images with camera poses and corresponding semantic labels (i.e., 180 images) from a randomly generated sequence of a certain scene. This fully-supervised setup acts as an upper bound on the semantic segmentation performance of Semantic-NeRF given abundant labelled training data.

However, in practice it is expensive and time-consuming to acquire accurate dense semantic annotations for all observed images in a scene.
Considering the redundancy in semantic labels among overlapping frames, we borrow the idea of key-framing from SLAM systems and hypothesise that providing labels for only selected frames should be enough to train the semantic representation efficiently.
We choose key-frames by evenly sampling from sequences and train the networks from scratch with semantic labels only coming from those selected key-frames, while the synthesis performance is always evaluated on all test frames.

\begin{figure}[!t]
\centering
\includegraphics[trim=0 3 0 0,clip,width=0.99\linewidth]{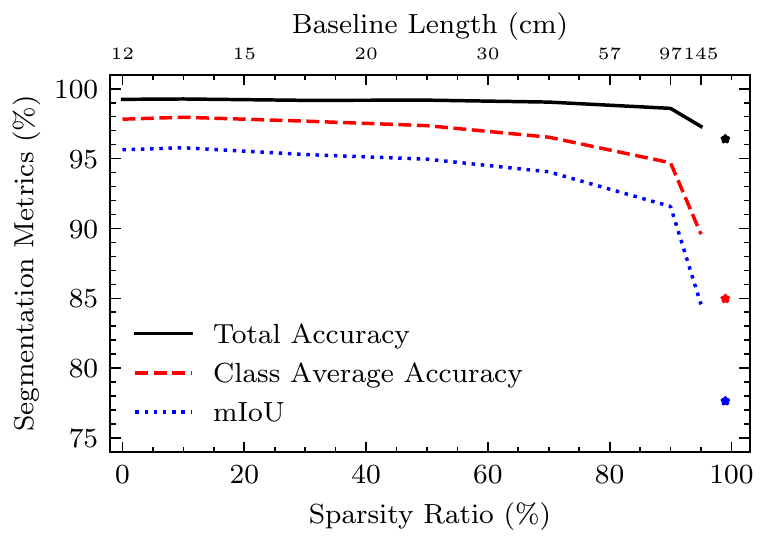}
\caption{Quantitative performance of Semantic-NeRF trained on Replica with sparse semantic labels. Sparsity ratio is the percentage of frames dropped compared to full sequence supervision. Three standard metrics are used to evaluate semantic segmentation performance on test poses (higher is better). Performance gracefully degrades with fewer labels due to uncovered or occluded regions, indicating the possibility of efficient dense labelling from fewer annotations. Results with only two labelled key-frames ($\star$) show remarkably competitive performance.
\label{fig:quanti-sparse-vs}}
    \vspace{2mm} \hrule \vspace{-7mm}
\end{figure}

Figure \ref{fig:quali-sparse-vs} and \ref{fig:quanti-sparse-vs} validate our assumption that semantics can be efficiently learned from sparse annotations with a sparsity ratio ranging from 0\% to 95\%, together with corresponding camera motion baselines as a complementary indication. Only marginal performance loss occurs when less than 10\% semantic frames are used, and this is mainly caused by renderings of regions which are unobserved or occluded from key-frames. To take this even further, we manually select just two key-frames (99\% sparsity ratio) from each scene to cover as much of the scene as possible. It turns out that our network, trained only with two labelled keyframes, can render accurate labels from various viewpoints. 

\vspace{-2mm}
\subsection{Semantic Fusion}
\vspace{-1.5mm}
In addition to being able to learn the semantic representation with sparse annotations due to the redundancy present in the semantic labels, another important property of Semantic-NeRF is that multi-view consistency between semantic labels is enforced.

In semantic mapping systems (e.g. \cite{Sunderhauf:etal:IROS2017, McCormac:etal:ICRA2017, Narita:etal:IROS2019}), multiple 2D semantic observations are integrated into a 3D map or target frames to produce a more consistent and accurate semantic segmentation.
Multi-view consistency is the key concept and motivation in semantic fusion, and the training process of Semantic-NeRF itself can be seen as a multi-view label fusion process. Given multiple noisy or partial semantic labels, the network can fuse them into a joint implicit 3D space so that we can extract a denoised label when we re-render the semantic labels from the learned representation back to input training frames.

We show the capability of Semantic-NeRF to perform multi-view semantic label fusion under a number of different scenarios: pixel-wise label noise, region-wise label noise, low-resolution dense or sparse labelling, partial labelling, and using the output of an imperfect CNN.

\vspace{-5mm}
\subsubsection*{Semantic Label Denoising}
\vspace{-2mm}
\paragraph*{Labels with Pixel-wise Noise}
We corrupt ground-truth training semantic labels by adding independent pixel-wise noise. Specifically, we randomly select a fixed portion of pixels per training frame and randomly flip their labels to arbitrary ones (including the void class). After training using only these noisy labels, we obtain denoised semantic labels by rendering back to the same training poses.

\begin{figure}[!t]
\centering
\includegraphics[width=0.97\linewidth]{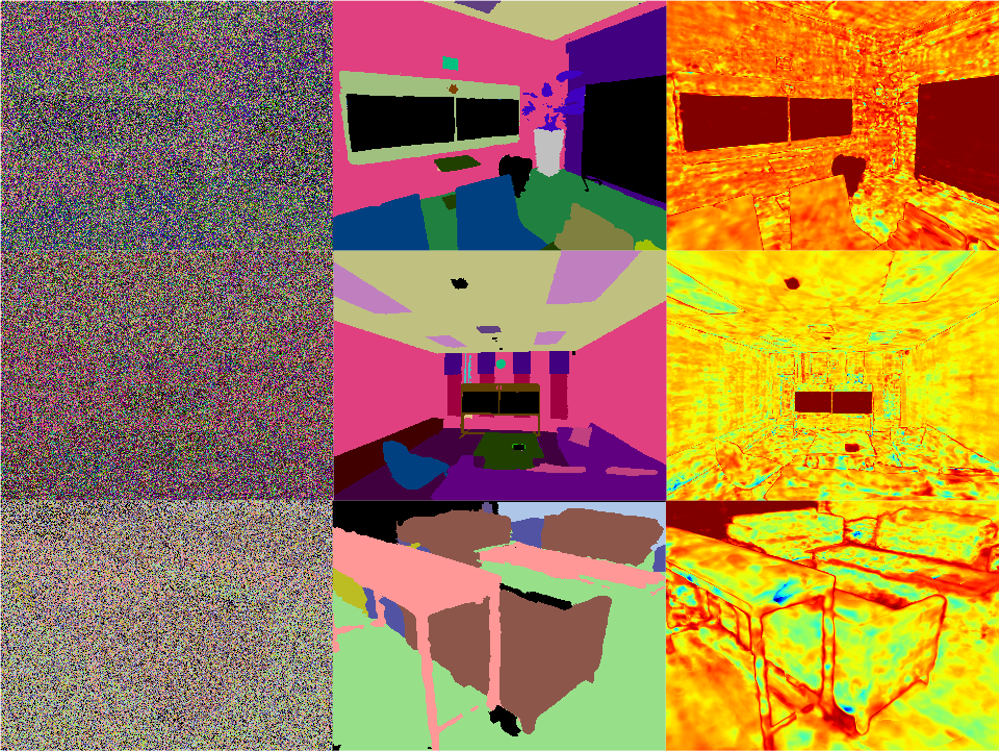}
\caption{Qualitative results for semantic denoising. Even when 90\% of all training labels are randomly corrupted, we can recover an accurate denoised semantic map. From left to right are noisy training labels, denoised labels rendered from the same poses after training,  and  information entropy. The overall high entropy we see in denoising tasks indicates the large inconsistency among noisy training labels. \label{fig:denoising}}
    \vspace{-3mm}
\end{figure}

Figure \ref{fig:denoising} shows qualitative results from label denoising. When 90\% of training  pixels are randomly flipped, and it is difficult even for a human to recognise the underlying structure of the scene, the denoised labels still retain accurate boundaries and detail, especially for fine structures. Compared with Figure \ref{fig:quali-sparse-vs}, the entropy in this denoising task is higher because the noisy training labels lack the multi-view consistency of clean ones. In addition, regions with void class tend to have the highest uncertainty since noisy pixels in void regions are not optimised during training. Quantitative results shown in Table \ref{tab:denoising} also confirm that accurate denoised labels are obtained after training-as-fusion.

While pixel-wise denoising with such severe corruption is not a realistic application, it is still a very challenging task and, more importantly, highlights our key observation that training itself is a fusion process which enables coherent renderings benefiting from the internal consistency of implicit joint representation.

\begin{table}
    \centering
    \resizebox{\linewidth}{!}{%
    \begin{tabular}{ccccc}
    \toprule 
    \multicolumn{2}{l}{{\bf Pixel-Wise Denoising}} & \multicolumn{3}{c}{{Metrics}}\\
         \cmidrule(l{3pt}r{3pt}){1-2} \cmidrule(l{3pt}r{3pt}){3-5} 
    \multicolumn{2}{l}{{Noise Ratio }} & {mIoU} & {Avg Acc} & {Total Acc}\\
     \cmidrule(l{3pt}r{3pt}){1-2} \cmidrule(l{3pt}r{3pt}){3-5} 
    \multirow{2}{*}{{50\% }} & \multicolumn{1}{l}{Input Label} & {0.191} & {0.534} & {0.533}\\
     & \multicolumn{1}{l}{Denoised Label} & {0.951} & {0.969} & {0.994}\\
    \midrule 
    \multirow{2}{*}{{90\% }} & \multicolumn{1}{l}{Input Label} & {0.041} & {0.145} & {0.145}\\
     & \multicolumn{1}{l}{Denoised Label} & {0.877} & {0.908} & {0.989}\\
    \bottomrule
    \end{tabular}
    }\\
    \resizebox{\linewidth}{!}{%
    \begin{tabular}{ccccc}
    \toprule 
    \multicolumn{2}{l}{{\bf Region-Wise Denoising}} & \multicolumn{3}{c}{mIoU} \\
    \cmidrule(l{3pt}r{3pt}){1-2} \cmidrule(l{3pt}r{3pt}){3-5} 
    \multicolumn{2}{l}{Noise Ratio} & { 30\%} & { 40\%} & { 50\%}\\
    \cmidrule(l{3pt}r{3pt}){1-2} \cmidrule(l{3pt}r{3pt}){3-5} 
    \multirow{2}{*}{{ Sort}} & \multicolumn{1}{l}{ Input Label} & {0.866} & {0.842} & {0.793}\\
     & \multicolumn{1}{l}{ Denoised Label} & {0.895} & {0.893} & {0.803}\\
    \midrule 
    \multirow{2}{*}{{ Even}} & \multicolumn{1}{l}{Input Label} & {0.741} & {0.692} & {0.684}\\
     & \multicolumn{1}{l}{ Denoised Label} & {0.796} & {0.747} & {0.733}\\
    \bottomrule
    \end{tabular}
    }
\caption{Quantitative evaluation for label denoising on Replica. Noise ratio is the percentage of changed pixels per frame, and for each instance the percentage of changed frames meeting selected criterion, respectively. mIoU is used for region denoising as it is more sensitive to the incorrect predictions on chair classes within the scene. Both tables are computed against clean training labels. \label{tab:denoising} }
\vspace{1mm} \hrule \vspace{-4mm}

\end{table}
\vspace{-3mm}
\begin{figure}[!t]
     \centering
     \includegraphics[width = 0.98\linewidth]{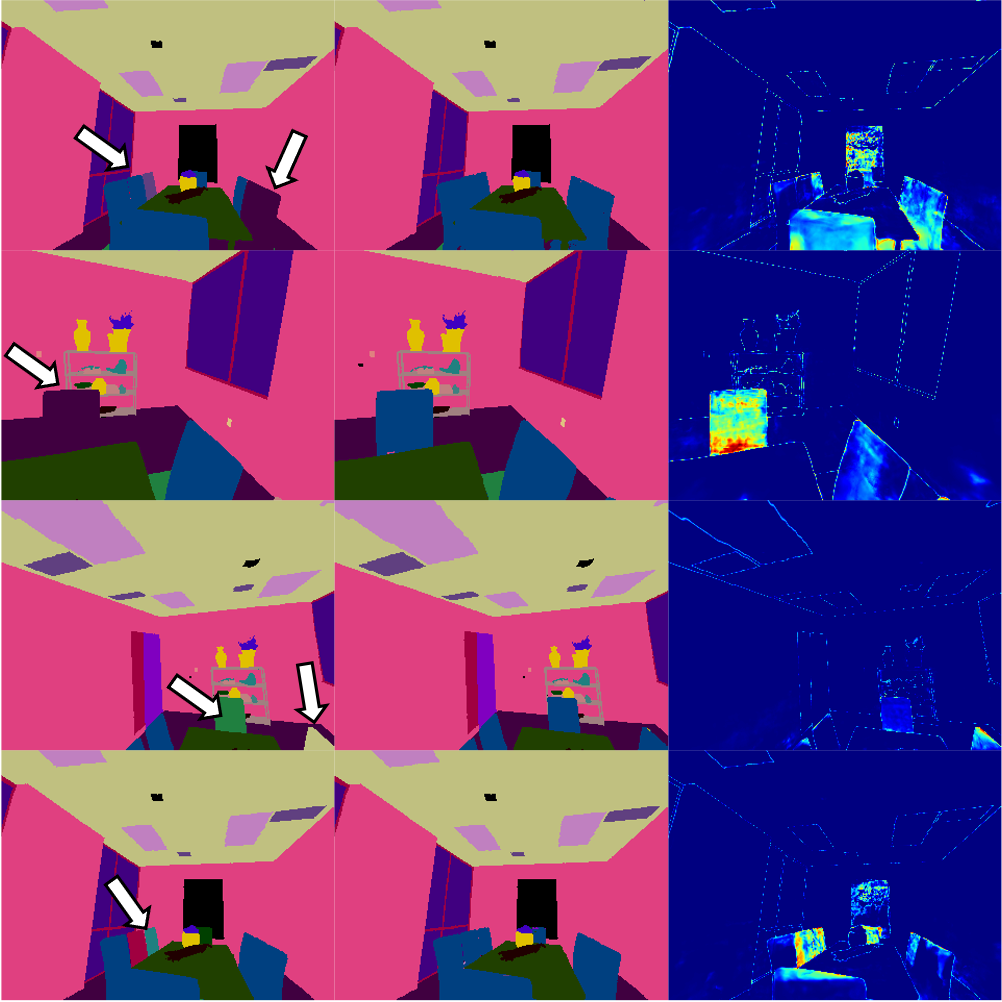}
    \caption{Qualitative results of rendered labels when we randomly change the training semantic class label ({\textcolor{blue}{blue}}) of chair instances. From left to right: training label with region-wise noise; recovered semantic labels rendered from the same poses; and  information entropy, highlighting regions with noisy predictions.} \label{fig:qualitative_rd}
    \vspace{1mm} \hrule 
    \vspace{-5mm}
\end{figure}

\vspace{2mm}
\paragraph*{Labels with Region-wise Noise}
We further validate the effectiveness of semantic consistency by randomly flipping the class labels of certain whole instances instead of pixels in the label maps. This is a better simulation of the behaviour of real single-view CNNs because a whole object can easily be labelled as a similar but incorrect class from an obstructed or ambiguous view.

We choose Replica Room\_2 containing 8 instances of chairs as the testing scene.
For each chair instance, we compute the occupied area ratio (i.e., ratio of the number of pixels belonging to that instance to the total number of pixels in the image for each ground truth label frame) and then sort the label maps in the sequence based on this occupied area ratio. Two criteria are used for selecting frames in which to randomly perturb each instance: (1) \textbf{Sort}: Select label maps with the least occupied area ratio. The intuition for this is that frames with partial observations are more likely to be mislabelled by semantic label prediction networks due to ambiguous context. (2) \textbf{Even}: Select label maps evenly from the sorted sequences introducing more large inconsistent regions into the training process. 

Figure \ref{fig:qualitative_rd} shows the qualitative results of the re-rendered semantic labels after training. We indeed observe that semantic labels of the chair instances can be corrected due to the enforcement of multi-view consistency during training. Table \ref{tab:denoising} also shows that there are steady improvements while it becomes much harder to render improved labels when a larger fraction of labels are perturbed.
\begin{figure*}[!t]
     \begin{subfigure}[b]{0.5\textwidth}
         \centering
         \includegraphics[width=0.95\linewidth, height=6cm]{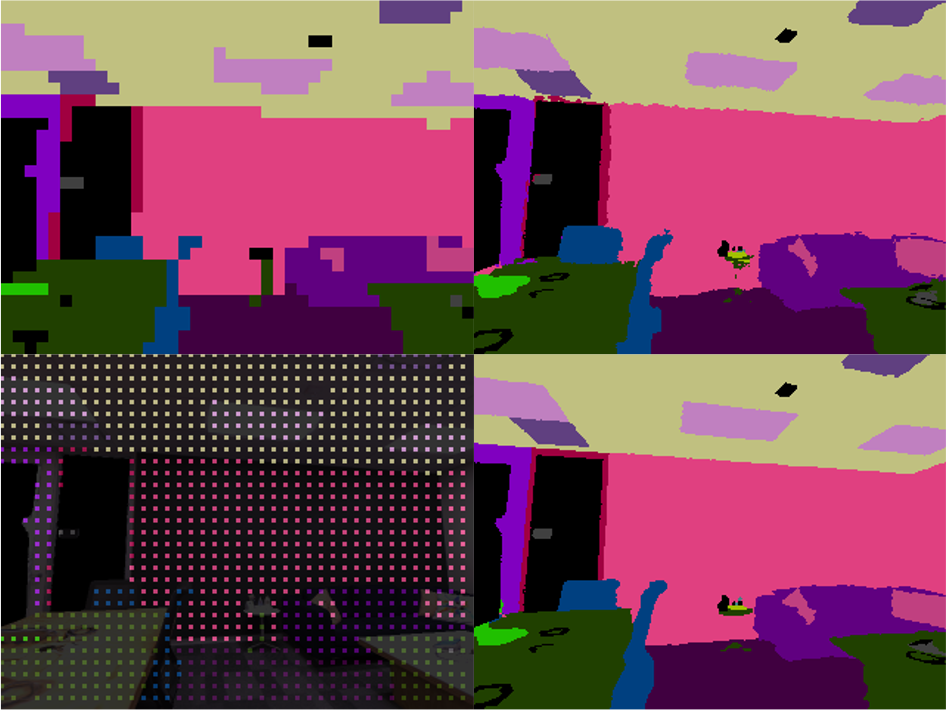}
         \vspace{-2mm}
         \caption{Super Resolution Scale ×8}
         \label{fig:spx8}
     \end{subfigure}
     \begin{subfigure}[b]{0.5\textwidth}
         \centering
         \includegraphics[width=0.95\linewidth, height=6cm]{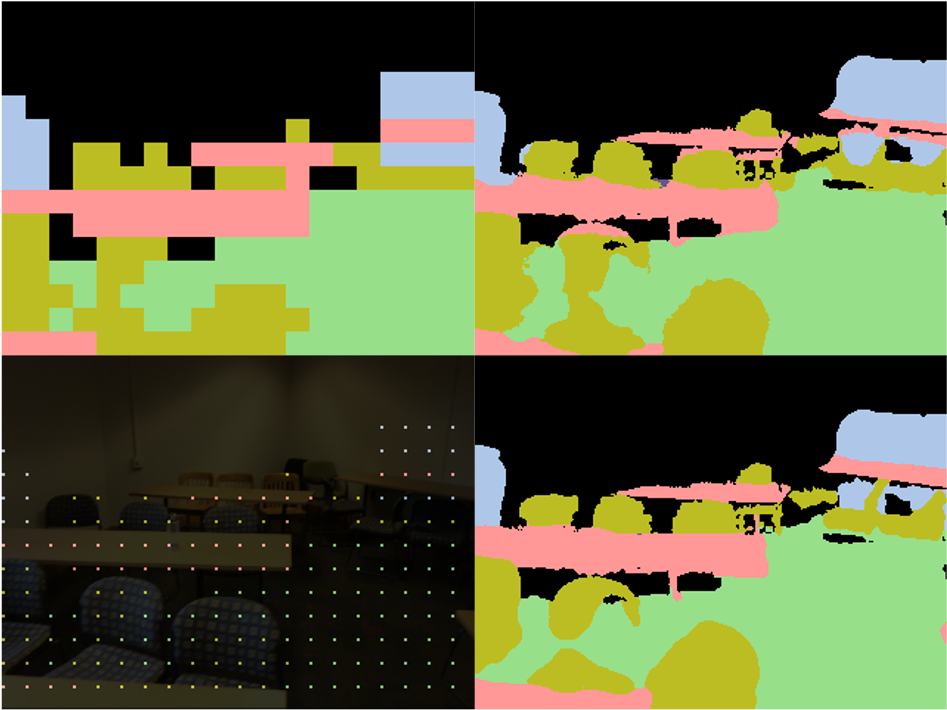}
        \vspace{-2mm}
         \caption{Super Resolution Scale ×16}
         \label{fig:spx16}
     \end{subfigure}
\vspace{-5mm}
\caption{Super-resolution: we train Semantic-NeRF with only low resolution labels (sparsely sampled or interpolated) and obtain super-resolved labels by re-rendering semantics from the same poses. Note that the sparse labels have been zoomed-in 4 times and overlaid on top of colour images for the ease of visualisation.
\label{fig:sp}}
 \vspace{1mm}
\hrule \vspace{-3mm}
\end{figure*}
\vspace{-3mm}
\subsubsection*{Super-Resolution}
\vspace{-1mm}
Semantic label super-resolution is a useful application for scene labelling as well. In an incremental real-time semantic mapping system, a light-weight CNN predicting low-resolution semantic labels might be adopted to reduce computational cost (e.g. \cite{Yoshikatsu:etal:IROS2018}).  Another possible use case is in a scene labelling tool, since manual annotation in coarse images is much more efficient.

Here we show that we can train Semantic-NeRF with only low-resolution semantic information but then render accurate super-resolved semantics for either the input viewpoints or novel views.
We test two different strategies to generate low-resolution training labels, with and without interpolation as shown in Figure~\ref{fig:sp}. Given a down-scaling factor $S=8$ for instance:
\vspace{-1mm}
\begingroup
\renewcommand\labelenumi{(\theenumi)}
\begin{enumerate}
    \setlength\itemsep{-1mm}
    \item  All ground truth labels are down-scaled from $\text{320}\times\text{240}$ to $\text{40}\times\text{30}$ before being up-scaled back to the original size using nearest neighbour interpolation.
    \item All pixels except those from the low-res label map (row and column divisible by $8$) are masked by the void class so as not to contribute to the training loss.
\end{enumerate}
\endgroup

\vspace{-2mm}
While method (1) uses interpolated labels to provide `dense' supervision to a sampled ray batch but will incorrectly interpolate some pixels, method (2) provides sparse but geometrically accurate labels. We report super-resolution performance on training poses from all Replica scenes with two scales $S=8$ and $S=16$ in Table ~\ref{tab: sp}. 
Figures \ref{fig:teaser} and \ref{fig:sp} show some examples where detailed semantic information is recovered through the fusion of many low-resolution or sparsely annotated semantic frames.

The promising results in semantic label denoising and super-resolution demonstrate one of the main advantages of jointly representing appearance, geometry and semantics: that missing or corrupted semantic labels in any one frame can be corrected through the fusion of many other frames.
We explore this property in more detail in the next section.
\vspace{-10mm}

\begin{center}
\begin{table}
\begin{centering}
\resizebox{0.98\linewidth}{!}{%
    \begin{tabular}{ccccc}
    \toprule 
    \multicolumn{2}{c}{{\bf Super-Resolution}} & \multicolumn{3}{c}{{Metrics}}\tabularnewline
    \cmidrule(l{3pt}r{3pt}){1-2} \cmidrule(l{3pt}r{3pt}){3-5} 
    \multicolumn{2}{c}{{Down-Scaling Factor}} & {mIoU} & {Avg Acc} & {Total Acc}\tabularnewline
    \cmidrule(l{3pt}r{3pt}){1-2} \cmidrule(l{3pt}r{3pt}){3-5} 
    \multirow{2}{*}{{Dense}} & {S=8} & {0.610} & {0.710} & {0.923}\tabularnewline
    \cmidrule{2-5}
     & {S=16} & {0.433} & {0.535} & {0.855}\tabularnewline
    \midrule 
    \multirow{2}{*}{{Sparse}} & {S=8} & {0.887} & {0.928} & {0.987}\tabularnewline
    \cmidrule{2-5} 
     & {S=16} & {0.800} & {0.866} & {0.977}\tabularnewline
    \bottomrule
    \end{tabular}}
\end{centering}
\caption{Quantitative evaluation of label super-resolution, with good performance with either sampled or interpolated low-resolution labels. The mIoU metric shows that sparse but geometrically accurate labels are more helpful for fine structures at high resolution.} \label{tab: sp}
\vspace{1mm} \hrule \vspace{-5mm}
\end{table}
\end{center}
\begin{figure*}[!t]
    \centering
    \includegraphics[width = 0.98\linewidth]{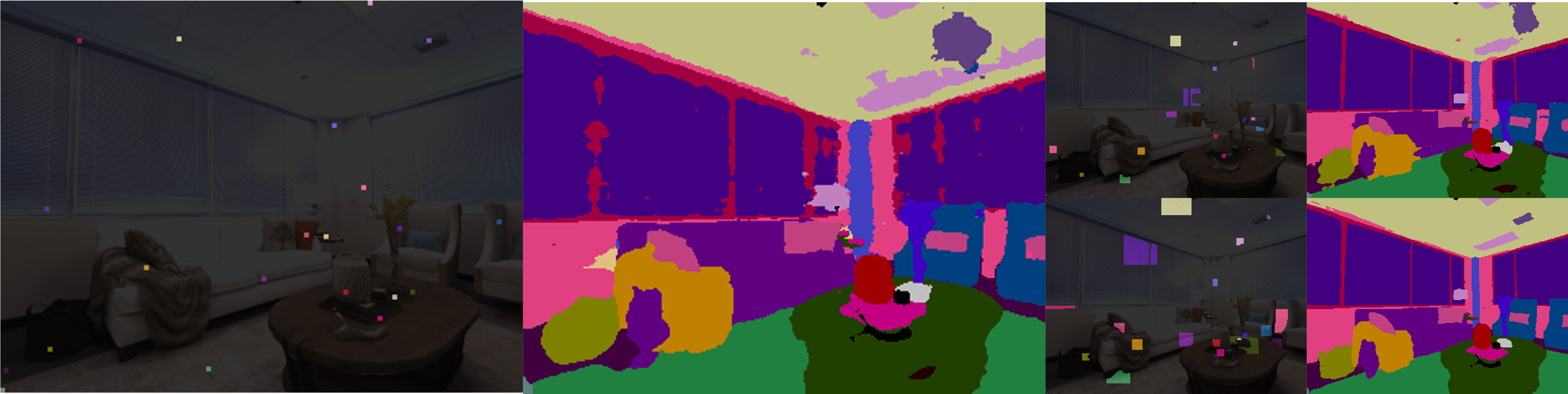}
    \caption{Label propagation results using partial annotations of a single-pixel, 1\% or 5\% of pixels per class within frames, respectively. Accurate labels can be achieved even from single-clicks, which are zoomed-in 9 times for visualisation purposes.} \label{fig:quantitative_partial}
    \vspace{-5mm}
\end{figure*}

\vspace{-2mm}
\subsubsection*{Label Propagation}
\vspace{-2mm}
Our super-resolution experiments have shown the ability of Semantic-NeRF to interpolate rich details from low-resolution annotations. For a practical scene-annotation tool, straightforward annotations from a user in the form of clicks or scratches are desirable, and expected that those sparse clicks can expand and propagate to accurately and densely label the scene.

To simulate user annotations, for each class within label maps we randomly select a continuous sub-region with which to apply a ground-truth label while leaving the rest unlabelled. Results in Figure \ref{fig:quantitative_partial} and Table \ref{tab:partial} show that supervision from one single pixel per class can lead to surprisingly high quality rendered labels with well preserved global and fine structure. Object boundaries are gradually refined when more supervision is available and the incremental improvements from more labels tend to saturate.

\begin{table}
\begin{centering}
\resizebox{0.98\linewidth}{!}{%
    \begin{tabular}{cccc}
    \toprule 
    {\bf Label Propagation} & \multicolumn{3}{c}{Metrics}\tabularnewline
    \cmidrule(l{3pt}r{3pt}){1-1} \cmidrule(l{3pt}r{3pt}){2-4} 
     \# Labelling per Class & mIoU & Avg Acc & Total Acc\tabularnewline
    \midrule 
    Single Click & 0.602 & 0.937 & 0.908\tabularnewline
    \midrule 
    1\% & 0.706  & 0.934 & 0.944\tabularnewline
    \midrule 
    5\% & 0.836  & 0.946 & 0.971\tabularnewline
    \midrule 
    10\% & 0.884 & 0.957 & 0.980\tabularnewline
    \bottomrule
    \end{tabular}
    }
\end{centering}
{\caption{Evaluation of label interpolation and propagation on Replica scenes using test poses. Even single-pixel supervision leads to competitive performance on the accuracy metrics, which highlights the effectiveness of the representation for interactive scene labelling.}\label{tab:partial}}
\vspace{1mm} \hrule \vspace{-5mm}
\end{table}

\vspace{-3mm}
\subsubsection*{Multi-view Semantic Fusion} \label{subsubsec: mvsf}
\vspace{-2mm}
We have shown that a semantic representation can be learned from sparse or noisy/partial supervisions. Here we further validate its practical value in multi-view semantic fusion using CNN predictions.

There have been several classical pixel-wise semantic fusion approaches \cite{Hermans:etal:ICRA2014,McCormac:etal:ICRA2017, McCormac:etal:3DV2018} to integrate monocular CNN predictions from multiple viewpoints to refine segmentation. For fair comparison, here we have separated out the multi-view fusion approaches from such systems. Two baseline techniques are: Bayesian fusion, where multi-class label probabilities are multiplied together and then re-normalised (e.g. \cite{McCormac:etal:ICRA2017}), and average fusion, which simply takes the average of all label distributions (e.g. \cite{McCormac:etal:3DV2018}).

To prepare training data in Replica dataset, we render two different sequences per Replica scene to cover various parts of scenes. Each sequence consists of 90 frames evenly sampled from 900 renderings of size $640 \times 480 $ with semantic labels remapped to NYUv2-13 class convention.
 
We choose DeepLabV3+ \cite{Chen:etal:ECCV2018} with a ResNet-101 backbone as the CNN model for monocular label predictions. To generate decent monocular CNN predictions and avoid over-fitting, we train DeepLab on SUN-RGBD \cite{Song:etal:CVPR2015}, and then fine-tune it using data from all Replica scenes except the one chosen for training Semantic-NeRF and label fusion evaluation. We repeat this fine-tuning process and train one individual DeepLab CNN model for each test scene.

Monocular CNN predictions of the test scene are used for two purposes: (1) training supervision for our scene-specific Semantic-NeRF model; (2) monocular predictions (per-pixel dense softmax probabilities) for baseline multi-view semantic fusion methods. We train Semantic-NeRF using posed colour images together with CNN-predicted labels for 200,000 steps and then re-render the fused semantic labels back to the training poses as fusion results.

It is important to note that both baseline fusion techniques require depth information to compute the dense correspondences between frames while ours only requires posed images. We report the average performance across all testing scenes in Table \ref{tab:mvsf}, in which ground truth depth maps are used for the two baseline approaches to represent a `best case scenario'. Our method achieves the highest improvement across all metrics, showing the effectiveness of our joint representation in label fusion.
\begin{center}
{\LARGE{}}
\begin{table}
\begin{centering}
\resizebox{0.98\linewidth}{!}{%
    \begin{threeparttable}
    \begin{tabular}{cccc}
    \toprule
    {\Large{}Semantic Fusion} & {\Large{}mIoU} & {\Large{}Avg Acc} & {\Large{}Total Acc}\tabularnewline
    \midrule 
    {\Large{}Monocular} & {\Large{}0.659} & {\Large{}0.763} & {\Large{}0.855}\tabularnewline
    \midrule 
    {\Large{}Bayesian Fusion {*}} & {\Large{}0.668} & {\Large{}0.764} & {\Large{}0.865}\tabularnewline
    \midrule 
    {\Large{}Average Fusion {*}} & {\Large{}0.586} & {\Large{}0.703} & {\Large{}0.814}\tabularnewline
    \midrule 
    {\Large{}Bayesian Fusion {\dag}} & {\Large{}0.666} & {\Large{}0.761} & {\Large{}0.862}\tabularnewline
    \midrule 
    {\Large{}Average Fusion {\dag}} & {\Large{}0.586} & {\Large{}0.708} & {\Large{}0.808}\tabularnewline
    \midrule 
    {\Large{}NeRF-Training (Ours)} & \textbf{\Large{}0.680} & \textbf{\Large{}0.772} & \textbf{\Large{}0.870}\tabularnewline
    \bottomrule
    \end{tabular}
    \begin{tablenotes}
    \item[*] \large{ Using ground truth depth for data association.}
    \item[\dag] \large{ Using learned depth of Semantic-NeRF for data association.}
    \end{tablenotes}
    \end{threeparttable}
    }
\end{centering}
{\caption{Comparison of multi-view semantic label fusion methods. Our approach relying on consistency of scene representations outperforms baselines aided with depth maps.}\label{tab:mvsf}} 
\vspace{1mm} 
\hrule \vspace{-5mm}
\end{table}
\end{center}

\vspace{-6mm}
\section{Conclusion and Future Work}
\vspace{-1.5mm}
We have shown that adding a semantic output to a scene-specific implicit MLP model of geometry and appearance means that complete and high resolution semantic labels can be generated for a scene when only partial, noisy or low-resolution semantic supervision is available. This method has practical uses in robotics or other applications where scene understanding is required in new scenes where only limited labelling is possible. 

An interesting direction for future research is interactive labelling, where the continually training network asks for the new labels which will most resolve semantic ambiguity for the whole scene.
\vspace{-2mm}
\section{Acknowledgements}
\vspace{-2mm}
Research presented in this paper has been supported by
Dyson Technology Ltd. Shuaifeng Zhi holds a China Scholarship Council-Imperial Scholarship. We are very grateful
to Edgar Sucar, Shikun Liu, Kentaro Wada, Binbin Xu and Sotiris Papatheodorou for fruitful discussions.

\typeout{}
{\small
\bibliographystyle{ieee_fullname}
\bibliography{robotvision}
}

\appendix
\twocolumn[{%
\begin{minipage}{\textwidth}
   \null
   \vspace*{0.375in}
   \begin{center}
      {\Large \bf Supplementary Material for\\In-Place Scene Labelling and Understanding with Implicit Scene Representation \par}
   \end{center}
   \vspace*{0.375in}
\end{minipage}
\vspace{-10mm}
\begin{center}
\centering
\includegraphics[width = 0.3\linewidth]{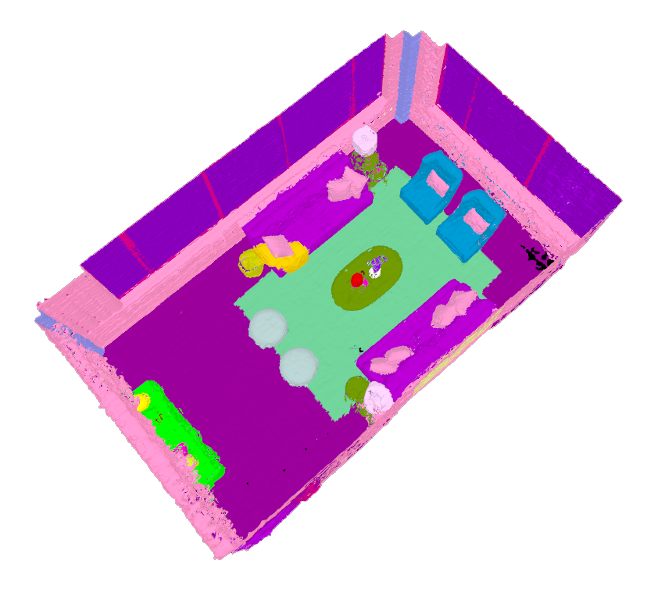}
\includegraphics[width = 0.3\linewidth]{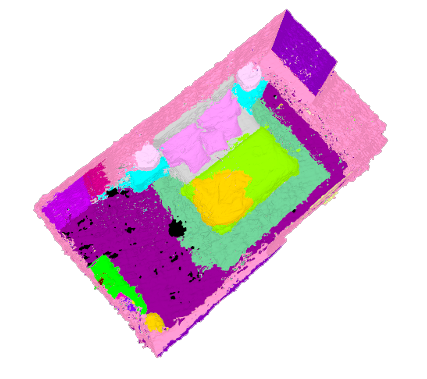}
\includegraphics[width = 0.33\linewidth]{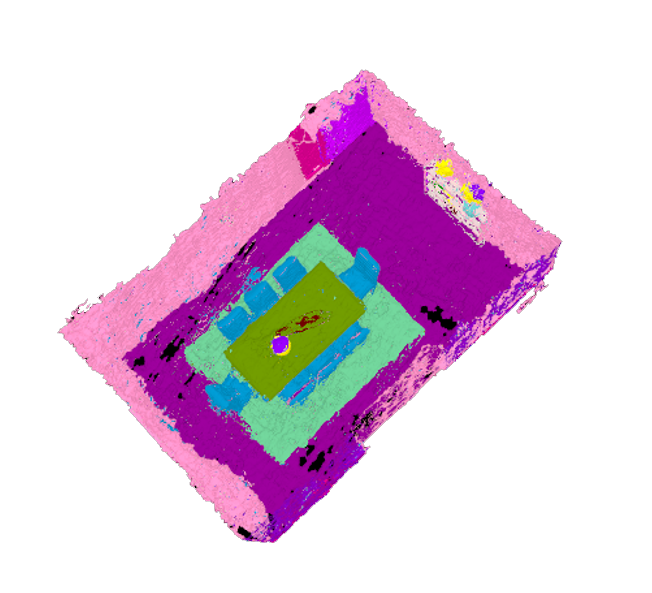}
\captionof{figure}{Semantic 3D reconstruction obtained using Semantic-NeRF. Note that our learned scene-specific 3D representation predicts decent geometry and semantics in occluded regions and fills the holes caused by unobserved regions to some extent.} \label{fig:sem_reco}
\vspace{2mm} \hrule
\end{center}}]

\section{Effects of Learning Semantics to Radiance and Geometry}
\vspace{-2mm}
Corresponding to Section 4.2 of the main paper, Table \ref{tab:quanti-synthesis} shows quantitative results for photometric and geometric reconstruction quality when projected to 2D on Replica scenes with and without semantics enabled. We observe no obvious difference between these two set-ups. Peak signal-to-noise ratio (PSNR) is used to measure the quality of the rendered colour images and the metrics used to evaluate the 2D depth maps are shown in Table \ref{tab:metric_defs}.

\begin{table}[!t]
\centering
\renewcommand{\arraystretch}{1.5}
\begin{tabular}{ll}
\hline
\multicolumn{2}{c}{2D Depth Metrics} \\
\hline
Abs Rel  & $\frac{1}{n}\sum{|d-d_{gt}|/d_{gt}}$ \\
Abs Diff & $\frac{1}{n}\sum{|d-d_{gt}|}$  \\
Sq Rel   & $\frac{1}{n}\sum{|d-d_{gt}|^2/d_{gt}}$ \\
RMSE     & $\sqrt{\frac{1}{n}\sum{|d-d_{gt}|^2}}$ \\
$\delta < 1.25^i$ & $\frac{1}{n}\sum{(\max{(\frac{d}{d_{gt}},\frac{d_{gt}}{d})} < 1.25^i)}$ \\
\hline
\end{tabular}
\caption{Definitions of depth metrics: $n$ is the number of valid depth pixels, $d$ and $d_{gt}$ are rendered depths at testing poses and ground truth depths, respectively.}
\label{tab:metric_defs}
\vspace{-5mm}
\end{table}

\vspace{-2mm}
\section{Semantic 3D Reconstruction from Posed Images}
After training semantic-NeRF with in-place annotation, we can also extract an explicit 3D scene from the learned MLP to inspect the implicit 3D representation.
Geometric meshes are extracted by first querying the MLP on dense 3D grids of the scene and then applying marching cubes. Attached semantic texture is rendered by treating the \textit{negative} normal direction of vertices in the mesh as the ray marching directions during volume rendering. 
We show qualitative results for three Replica room scenes in Figure \ref{fig:sem_reco}.

\vspace{-3mm}
\section{Network Architecture}
\vspace{-2mm}
Axis-aligned positional encoding (PE) of 3D positions are fed to both first and intermediate fully-connected (FC) layers with 256 neurons and ReLU activations before predicting volume density. Additional FC layers with 128 neurons are used for \textit{view-invariant} semantics and \textit{view-dependent} radiance after merging input viewing directions.

The length of positional encoding $L$ relates to the maximum frequency used and affects the rendering quality. In label propagation task, we find that using only low-frequency components ($L=5$) leads to over-smoothed 2D renderings, while using high-frequency ones ($L=40$) leads to noisy interpolations, which aligns with findings in recent literature [\textcolor{green}{16}, \textcolor{green}{30}]. $L$ of 10 empirically performs the best. 

\vspace{-3mm}
\section{More Qualitative Results}
\begin{figure*}
\centering
\includegraphics[width = 1\linewidth]{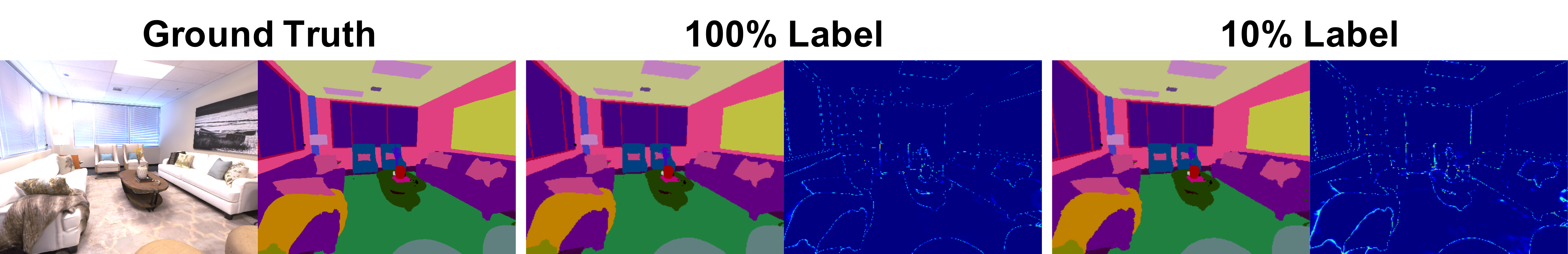}

\vspace{5mm}

\includegraphics[width = 1\linewidth]{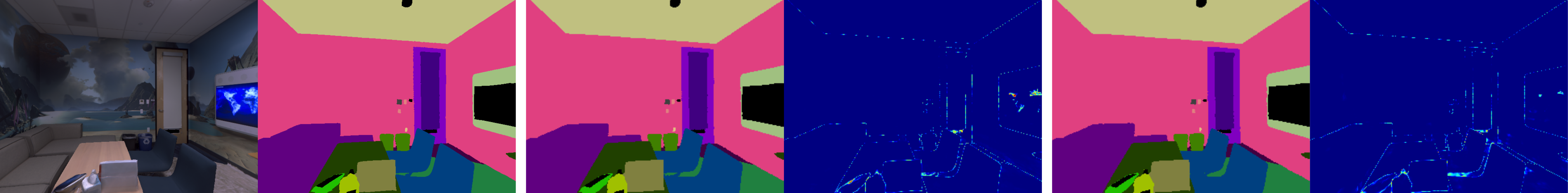}

\vspace{5mm}

\includegraphics[width = 1\linewidth]{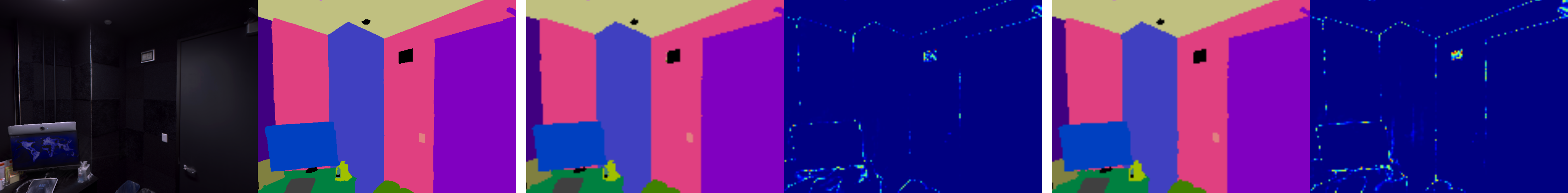}

\vspace{5mm}

\includegraphics[width = 1\linewidth]{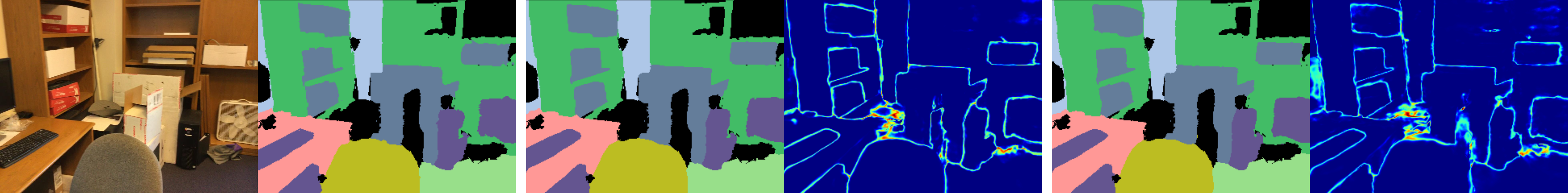}

\vspace{5mm}

\includegraphics[width = 1\linewidth]{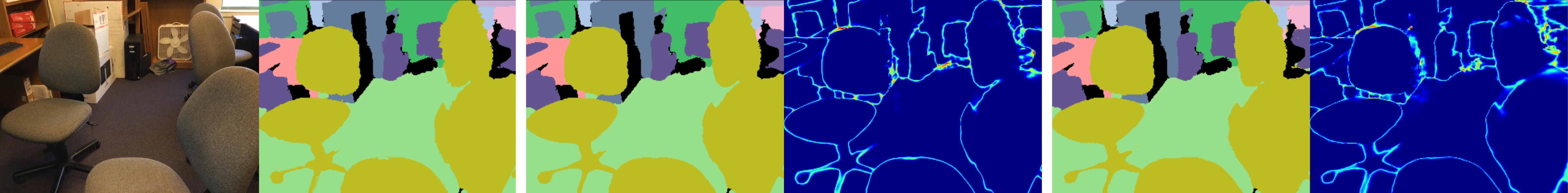}

\vspace{5mm}

\includegraphics[width = 1\linewidth]{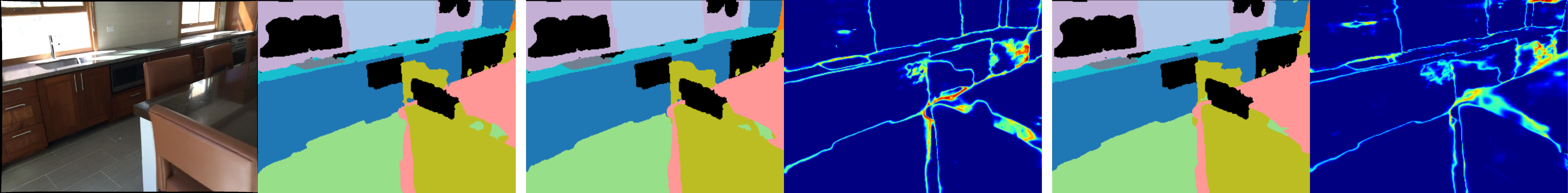}

\vspace{5mm}

\includegraphics[width = 1\linewidth]{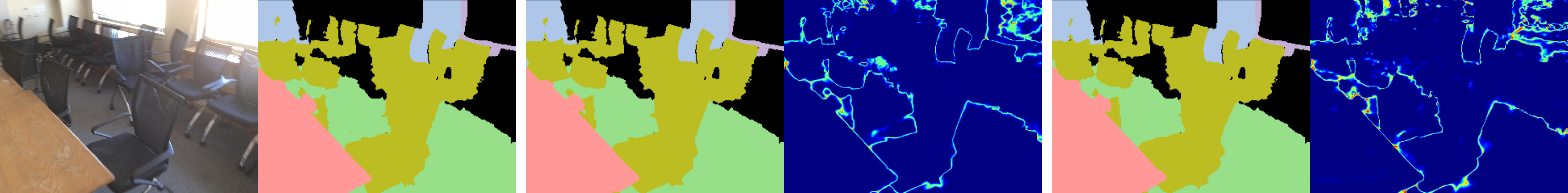}

\caption{View-synthesis results. } \label{fig:supple_vs}
\end{figure*}

\begin{figure*}
\includegraphics[width = 1\linewidth]{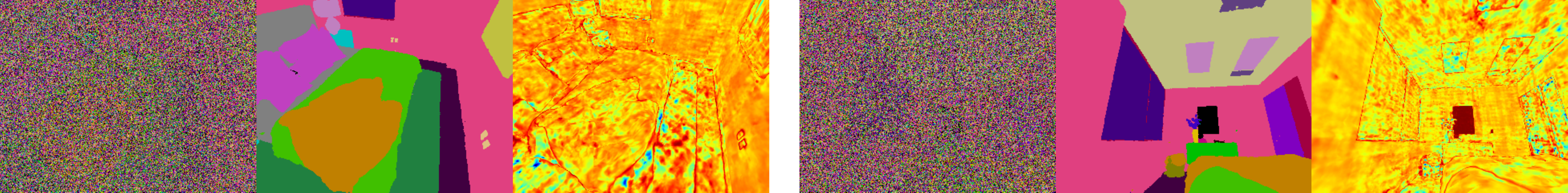}

\vspace{5mm}

\includegraphics[width = 1\linewidth]{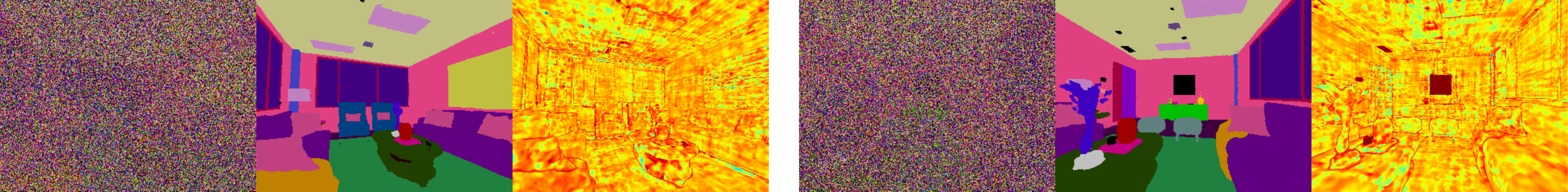}

\vspace{5mm}

\includegraphics[width = 1\linewidth]{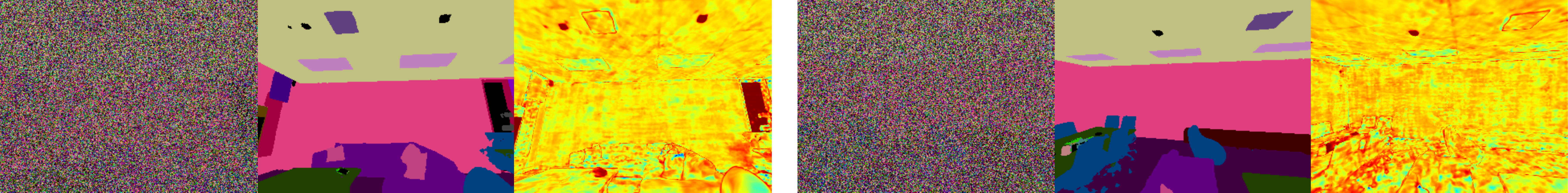}

\vspace{5mm}

\includegraphics[width = 1\linewidth]{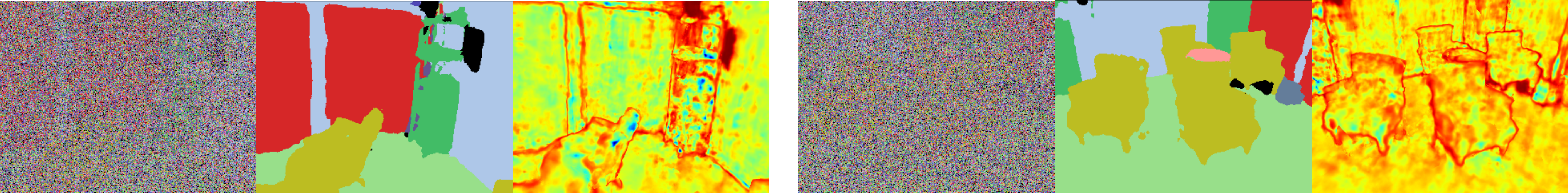}

\vspace{5mm}

\includegraphics[width = 1\linewidth]{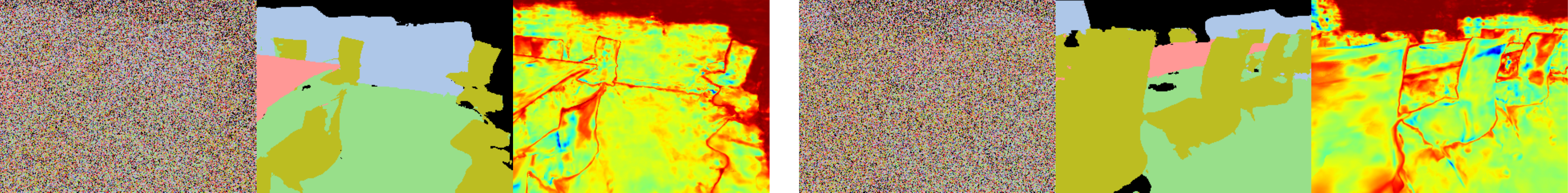}

\caption{Pixel-wise denoising of semantic labels with 90\% noise ratio. } \label{fig:supple_pd}

\end{figure*}

\begin{figure*}
    \begin{subfigure}[b]{0.5\textwidth}
        \centering
        \includegraphics[width = 1\linewidth]{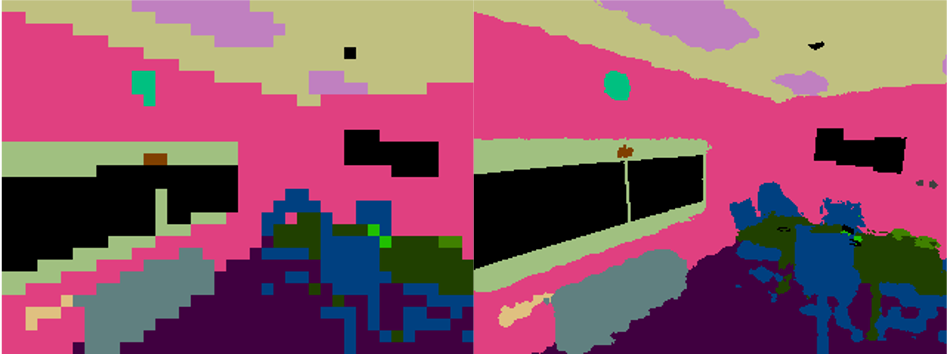}
        
        \vspace{2mm}
        
        \includegraphics[width = 1\linewidth]{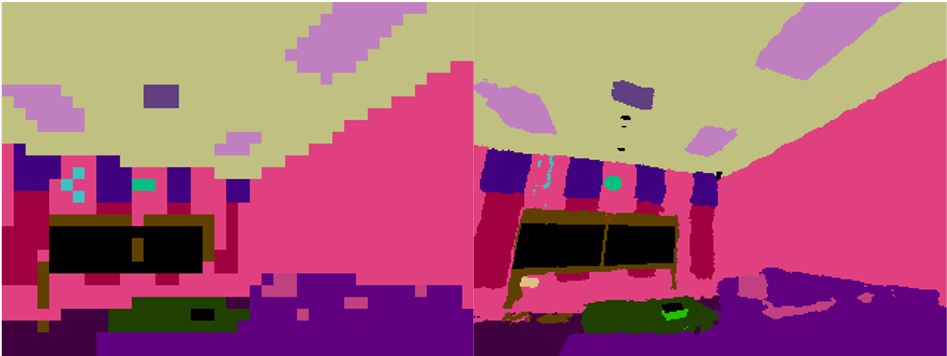}
        
        \vspace{2mm}
        
        \includegraphics[width = 1\linewidth]{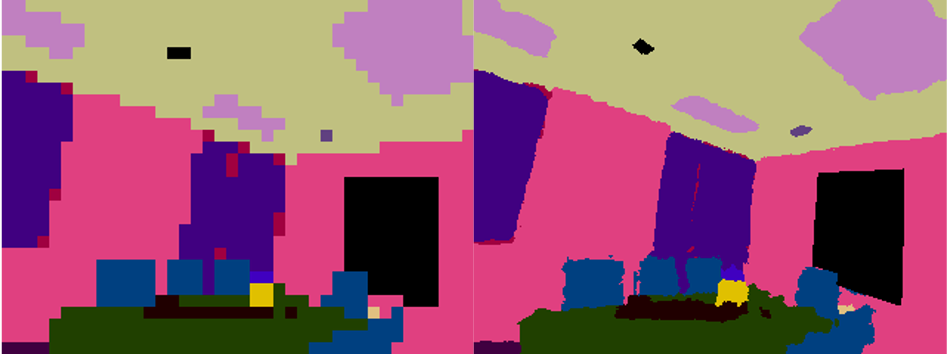}
        
        \vspace{2mm}
        
        \includegraphics[width = 1\linewidth]{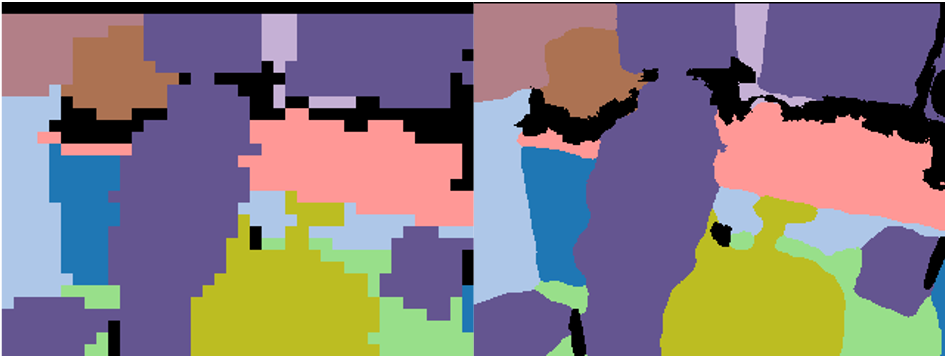}
        
        \vspace{2mm}
        
        \includegraphics[width = 1\linewidth]{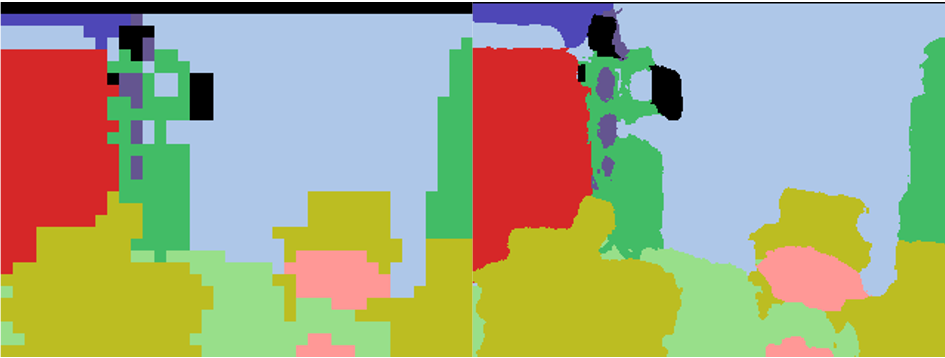}
        
        \vspace{2mm}
        
        \includegraphics[width = 1\linewidth]{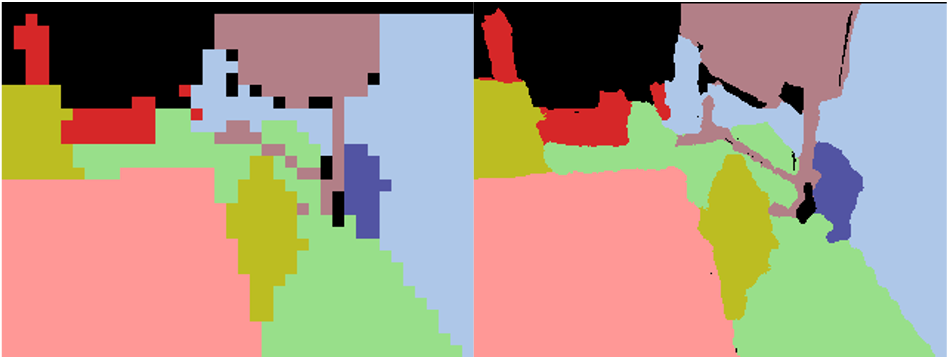}

        \vspace{-2mm}
        \caption{Super-resolution using coarse label}
    \end{subfigure}
    \begin{subfigure}[b]{0.5\textwidth}
        \centering
        \includegraphics[width = 1\linewidth]{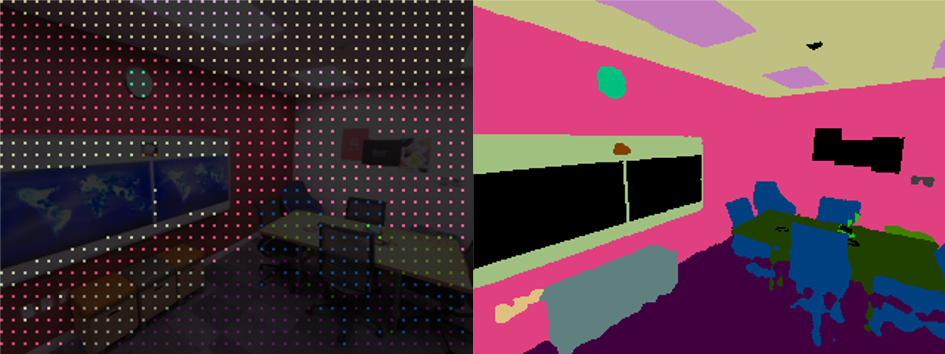}
        
        \vspace{2mm}
        
        \includegraphics[width = 1\linewidth]{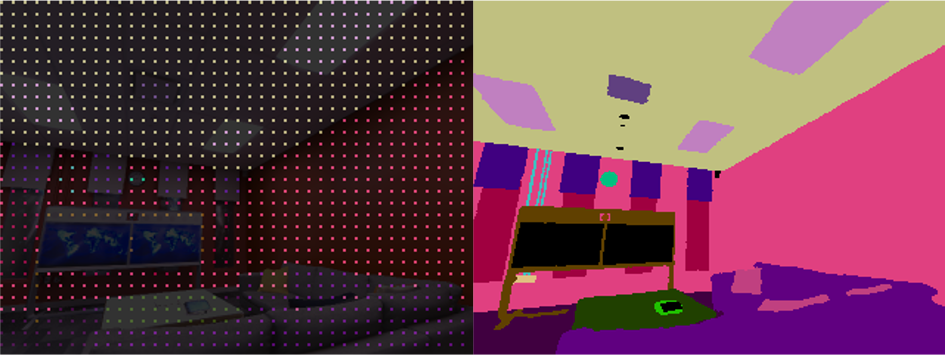}
        
        \vspace{2mm}
        
        \includegraphics[width = 1\linewidth]{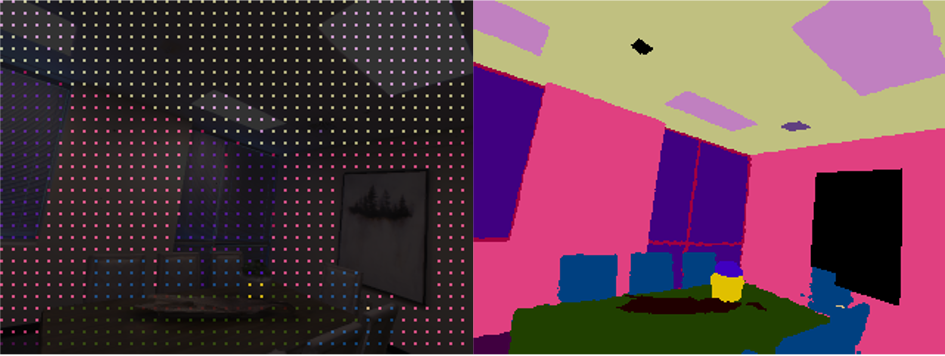}
        
        \vspace{2mm}
        
        \includegraphics[width = 1\linewidth]{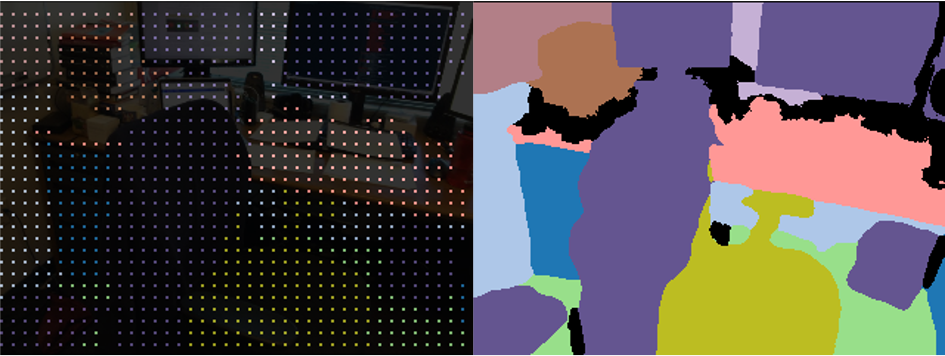}
        
        \vspace{2mm}
        
        \includegraphics[width = 1\linewidth]{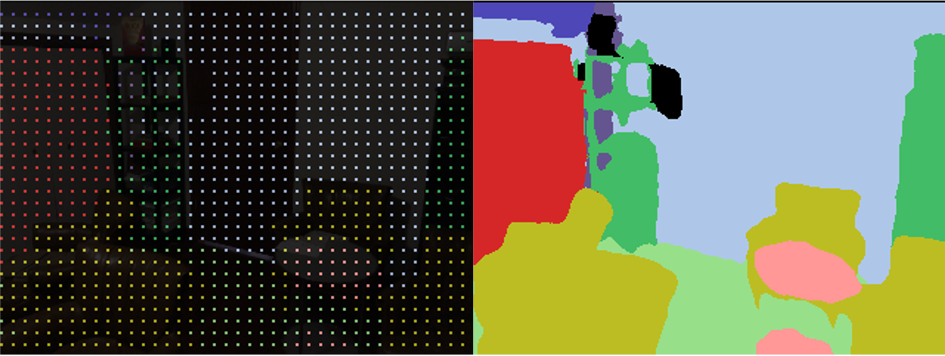}
        
        \vspace{2mm}
        
        \includegraphics[width = 1\linewidth]{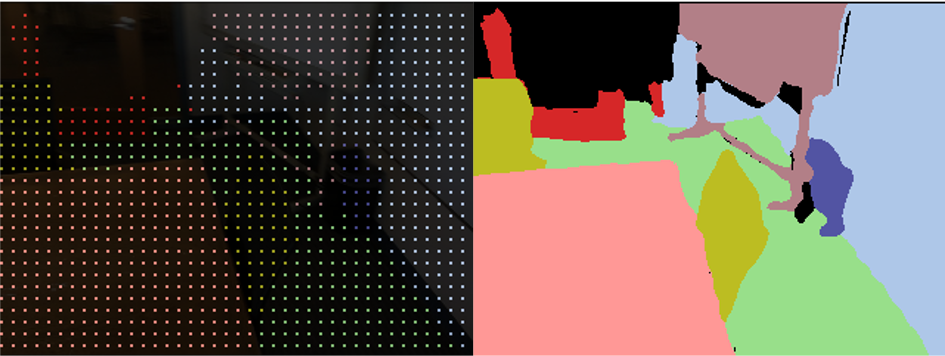}

        \vspace{-2mm}
        \caption{Super-resolution using sparse label}
    \end{subfigure}
\caption{Label super-resolution (×8) results.} \label{fig:supple_sp}
\end{figure*}

\vspace{-1mm}
Here we show more examples of qualitative results in Figure \ref{fig:supple_vs}, \ref{fig:supple_pd}, \ref{fig:supple_sp} for semantic view synthesis, label denoising and super-resolution, respectively.

We kindly urge readers to watch our \textit{supplementary video} on project page \url{https://shuaifengzhi.com/Semantic-NeRF} which highlights the accuracy and
consistency of semantic renderings in various situations and applications.

\begin{table*}[!t]
\begin{center}
\centering
\resizebox{0.95\linewidth}{!}{%
    \begin{tabular}{ccccccccc}
    \toprule 
    \multirow{2}{*}{{Network Set-up}} & \multicolumn{7}{c}{{Depth}} & {RGB}\tabularnewline
    \cmidrule{2-9} \cmidrule{3-9} \cmidrule{4-9} \cmidrule{5-9} \cmidrule{6-9} \cmidrule{7-9} \cmidrule{8-9} \cmidrule{9-9} 
     & {AbsRel\textdownarrow{}} & {AbsDiff\textdownarrow{}} & {SqRel\textdownarrow{}} & {RMSE\textdownarrow{}} & {$\delta<1.25$\textuparrow{}} & {$\delta<(1.25)^{2}$\textuparrow{}} & {$\delta<(1.25)^{3}$\textuparrow{}} & {PSNR\textuparrow{}}\tabularnewline
    \midrule 
    \multirow{1}{*}{{W/ Semantics}} & {0.017} & {0.032} & {0.007} & {0.096} & {0.993} & {0.997} & {0.998} & {32.27}\tabularnewline
    \midrule 
    {W/O Semantics} & {0.018} & {0.032 } & {0.009} & {0.102} & {0.993} & {0.996} & {0.998} & {32.80}\tabularnewline
    \bottomrule
    \end{tabular}}
\end{center}
\vspace{-5mm}
\captionof{table}{Quantitative evaluation of effects of predicting semantics on appearance and geometry on Replica dataset. }
\label{tab:quanti-synthesis}
\vspace{2mm}
\hrule
\vspace{5mm}
\end{table*}

\end{document}